\documentclass[10pt]{article}

\usepackage{lineno,hyperref}
\modulolinenumbers[5]

\usepackage{wrapfig}
\usepackage{listings}
\usepackage{amssymb}
\usepackage{amsmath}
\usepackage{esint}

\usepackage{amsmath, amssymb}
\usepackage[all]{xy}
\usepackage{graphicx}

\usepackage{physics}

\begin{document}


\title{Distinction Graphs and Graphtropy:  \\ A Formalized Phenomenological Layer \\
Underlying Classical and Quantum Entropy, \\
Observational Semantics \\
and Cognitive Computation }

\author{Ben Goertzel}





\maketitle

\begin{abstract}
A new conceptual foundation for the notion of ``information'' is proposed, based on the concept of a  {\it distinction graph}: a  graph in which two nodes are connected iff they cannot be distinguished by a  particular observer.   The ``graphtropy'' of a  distinction graph is defined as the average connection probability of two nodes; in the case where the distinction graph is a  composed of disconnected components that are fully connected subgraphs, this is equivalent to Ellerman's logical entropy, which has straightforward relationships to Shannon entropy.   Probabilistic distinction graphs and probabilistic graphtropy are also considered, as well as connections between graphtropy and thermodynamic and quantum entropy.   

The semantics of the Second Law of Thermodynamics and the Maximum Entropy Production Principle are unfolded in a  novel way, via analysis of the cognitive processes underlying the making of distinction graphs.   The Second Law is re-cast in the form: ``The graphtropy of an observer's memory graph will never decrease," which is argued to hold true under specific assumptions regarding the observer's cognitive structure.   This evokes a  connection between graphtropy and consciousness theory, in which complex intelligence is seen to correspond to states of consciousness with intermediate graphtropy, which are associated with memory imperfections that violate the assumptions leading to derivation of the Second Law.   

A connection with algorithmic information theory is made: In the case where nodes of a  distinction graph are labeled by computable entities, graphtropy is shown to be monotonically related to the average algorithmic information of the nodes (relative to to the algorithmic information of the observer).  

A quantum-mechanical version of distinction graphs is considered, in which distinctions can exist in a  superposed state; this yields to graphtropy as a  measure of the impurity of a  mixed state, and to a  concept of ``quangraphtropy'' that connects Feynman-type sums with graphtropy of distinction graphs.  

A novel computational model called Dynamic Distinction Graphs (DDGs) is formulated, via enhancing distinction graphs with additional links expressing causal implications.   DDGs enable a detailed model of the ``observers'' that play a central role in distinction graph theory and graphtropy. 

Via these diverse considerations, the position is advanced that founding information in the graph of distinctions perceivable by a  particular observer (who is himself a set of distinctions and causal relations therebetween), yields insight into many of the  multiple aspects of physical and mental reality in which the ``information" concept plays a  role.
\end{abstract}



\tableofcontents

\section{Introduction}

``Information'' is increasingly accepted as a  fundamental concept; it is now somewhat commonly posited that the physical world \cite{lloyd2006programming} and the thinking mind \cite{bateson1979mind} are both composed of information.

The basic nature of information, however, is not made especially clear in traditional mathematical, physical or engineering presentations of information theory and related ideas.

Often ``information'' is grounded in terms of ``entropy''.   Entropy has a  clear meaning in thermodynamics, and is a  fundamental concept in numerous other areas of science including quantum mechanics, communication theory, statistics, computational linguistics, AI and others.  Related concepts such as the Maximum Entropy Principle have a  very broad applicability as well.   However, the foundations of the entropy concept are not as plain as one might think.   For instance the thermodynamic and communication-theoretic notions of entropy are often discussed as if they were interchangeable, but on the mathematical level are interrelated only by an approximation.  The role of the observer in entropy-related phenomena such as the Second Law of Thermodynamics is also less clear than is generally suggested, and is rarely grounded in a  serious model of the ``observer'' as a  particular system in the world.

Here we seek to establish a  new foundation for the information concept, via taking a  slightly different approach both  mathematically and conceptually.

Mathematically, we begin with Ellerman's notion of ``logical entropy" \cite{ellerman2013introduction} and extend it to a  new concept of ``graph entropy" or {\it graphtropy}.    Graphtropy is an extremely simple quantity -- merely the ``mean connection weight" in a  graph.  However, this simple measure turns out to have an intriguing semantic interpretation, if one focuses on ``distinction graphs," in which the existence or weight of a  link between two nodes $a$ and $b$ denotes the propensity of a  certain observing system to confuse $a$ with $b$. 

Conceptually, our key innovation is to take the observer seriously.   To understand information, we suggest, one should begin with a  specific observer, and ask what distinctions this observer is capable of making.   This leads to a  distinction graph and to graphtropy as one among a  number of interesting quantities to compute. 

The answer to the question ``What is information?", is then posited to be: 

\begin{itemize}
\item a  body of information is a  set of distinctions, drawn by some particular observer
\item There are various ways of quantifying the {\it amount} of information in a  certain {\it body} of information (i.e. in a  certain set of distinctions).   The graphtropy (the mean amount of distinctiveness between entities) is one key measure of this sort.
\end{itemize}

\noindent Logical entropy, with clear relationships to traditional Shannon entropy, emerges as the formula for the graphtropy of a  body of information if one makes certain special assumptions about the structure of this body.   These assumptions are not accurate for real observers, but are often good enough approximations for practical purposes.

On a  philosophical level, the perspective given here is inspired somewhat by the thinking of G. Spencer-Brown \cite{brown1969laws} and Lou Kauffman \cite{kauffman1987self}, who have emphasized the foundational role of the concept of {\it distinction} in explaining how an observer creates a  world, and how a  world creates an observer.   However, the types of mathematical analysis pursued here are quite different from (and complementary to) what these two authors have done.   There is also a  very close connection between the ideas given here and Isaacson and Kauffman's notion of ``recursive distinctioning"  (RD) \cite{isaacson2016recursive}.     RD is concerned with constructing graphs whose edges represent distinctions, and then studying these graphs.   However, the specific work presented so far under the RD umbrella involves distinction graphs with rigid geometric structure (e.g. square lattices), and the algebras implicit in what happens when one labels these graphs in certain ways; interesting considerations, but quite different than the directions we pursue here (though probably there are relationships to be unfolded).

Graphtropy can be applied, as a  mathematical measure, independently of the interpretation of graphs as distinction graphs.   However, it is the use of graphtropy in the context of distinction graphs that will preoccupy us here the most.   From this perspective, graphtropy allows the exploration of entropy and related concepts at a  level more primitive and foundational than the analyses associated with the standard entropy formulae.   One begins with the ``observer" as a  system that has a  certain capability to distinguish pairs of entities from each other, and builds a  distinction graph based on a  specific observer's capability of distinction, and then studies quantities associated with this distinction graph -- thus obtaining a  new view of traditional information-theoretic formulae, and also some additional distinctive quantities of interest.

Graphtropy can be defined and calculated in a  quantum-mechanical context.   If one has distinctions that are ``superposed between being drawn and not being drawn," in the way that quantum logic allows, then one can calculate graphtropy in a  way that takes this into account, still getting a  meaningful quantitative measure of the amount of information in a  certain body of information.   The study of graphtropy on complexly structured distinction graphs may potentially shed light on the nature of weak quantum coherence.

Graphtropy connects with thermodynamics, in different (though related) ways to the links between these areas and traditional entropy.  Both the maximum entropy principle and the maximum entropy production principle appear meaningful in a  graphtropic context, though with considerable mathematical complexities that are mostly indicated rather than resolved here.   The Second Law of Thermodynamics can be studied in a  novel way using distinction graphs, and in this context takes the form of a  statement about the subjective perspectives of certain classes of observers.   If one assumes an observer's memory retains data about distinctions observed between states of a  system, then one arrives at the conclusion that {\it the graphtropy of the observer's memory graph will never decrease}, which is a  well-founded cognitive version of the Second Law.

Of course, a  real-world system's memory is going to be strictly bounded relative to the totality of its historical perceptions, and so the assumptions underlying the graphtropic Second Law will not hold.   This ties in with the observation that real-world intelligent systems tend to occupy conditions of intermediate graphtropy, corresponding to states of consciousness inbetween the purely ``oceanic'' state of minimum graphtropy and the ultimately ``individuating'' state of maximal graphtropy.

Graphtropy also connects with algorithmic information theory.   If the nodes in a  distinction graph are associated with entities possessing various degrees of complexity (in the sense of algorithmic information), then the graphtropy emerges as a  nonlinear, monotonic transform of the mean degree of complexity of these entities, as compared to the complexity of the observer.    Graphtropy is thus, in a  sense, a  monotonic transform of the algorithmic information of the observed relative to the observer.   This connects computing theory with more philosophically basic considerations related to bodies of perceived distinctions.

Finally we articulate a  simple, novel computing model based on distinction graphs, obtained via augmenting the basic distinction graph model with causal implication links to yield what are called Dynamic Distinction Graphs (DDGs).   In this model the complexity of a  node has to do with the set of computational operations that are embodied by the graph links connecting that node.   The goal here is to show how computing can emerge from very simple phenomenological entities.   Such a  model may be useful for understanding computing at the border between classical and quantum, and at the border between physical and cognitive. It also gives an in-depth formal model for the ``observers'' that play such a critical role in distinction graph theory.   A distinction graph exists relative to an observer, so it's elegant that an observer can in turn be modeled as a dynamic distinction graph; and a time-series of distinction graphs can be data-mined to form a DDG which can in turn be considered as an observer.

This paper is an initial foray into a  new area, and much remains to be done.   However, I believe enough is presented here to make the key point.   If one begins by taking the observer seriously, and constructing a  graph representing the distinctions this observer is capable of making, then one can calculate simple quantities associated with this graph and arrive at a  more foundational understanding of ``what is information, relative to an observer."   Traditional notions of entropy then emerge from these same graphs under special structural assumptions.   Traditional notions of entropy are extremely useful in various applications, but we believe that the study of distinction graphs provides a  deeper view of the concept of ``information" and thus of the foundations of information theory, and a  pointer into a  rich domain of associated, important and minimally explored ideas.

\section{Logical Entropy}

In this section we briefly summarize some mathematical and conceptual preliminaries: Ellerman's notion of {\it logical entropy}, which connects elegantly to traditional Shannon entropy, and which also turns out to be a  special case of our concept of graphtropy.

Given a  partition $\pi = \{B_1, \ldots, B_n\}$ of a  set $U$, David Ellerman \cite{ellerman2013introduction}  defines a  {\it dit} or {\it distinction} of the partition as an ordered pair $(u,u') \in u \times U$ so that $u$ and $u'$ are in different elements of the partition.   The set of distinctions of the partition $\pi$ is the dit set $\textrm{dit}(\pi)$.   The {\it logical entropy} of the partition is then defined as

$$
h(\pi) =  \frac { |\textrm{dit}({\pi})|} {| u \times u |}
$$

\noindent If one sets $p_i = \frac{B_i}{B}$, then it follows that

$$
h(\pi) = 1 - \sum_{i=1}^n p_i^2
$$

\subsection{Quantum Logical Entropy}

The quantum variant of these ideas follows similarly \cite{ellerman2016classical}.   Let $V$ be an $n$-dimensional Hilbert space $\mathbb{C}^n$.   Let $\ket{\Phi_i}, i=1\ldots m$ be a  set of orthornormal vectors in $V$, and let $p=(p_1, \ldots, p_m)$ be a  probability distribution; then

$$
\rho(\Phi) = \sum_{i=1}^m p_i \ket{Phi_i} \bra{Phi_i}
$$

\noindent is a  density matrix, and any positive operator on $V$ of trace $1$ can be represented this way.   The {\it quantum logical entropy} is then defined as

$$
h(\rho) = 1 - tr[\rho^2]
$$

\noindent -- a  formula often referred to as the "purity'' of the density matrix.

\section{Graphtropy}

Ellerman's logical entropy  may be viewed as a  special case of a  more general concept of {\it graph logical entropy}, or more concisely ``graphtropy."   Given $N$ elements viewed as nodes of a  graph, a  partition of the elements corresponds to a  ``partition graph" that consists of a  set of disjoint completely-connected subgraphs spanning the $N$ elements.   We will describe here a  way of calculating a  ``logical entropy" for a  general graph, which reduces to Ellerman's logical entropy for the special case of a  partition graph.  After presenting the formal notion of graphtropy we will outline a  natural semantic interpretation in terms of the distinctions made by an observer.

Given an undirected graph $G = (U_G, E_G)=(U,E)$, we may define a  ``dit'' as a  pair $(u, u')$  of nodes that don't have an edge between them in the edge-set $E$.   In the case that G is a  partition graph, the dit set $\textrm{git}(G)$ of G is the dit set of the corresponding partition.

We then define the {\it graphtropy} of $G$ as

$$
h(G) = \frac{ |\textrm{git(G)}|}{ |U_G|^2}
$$

As Ellerman does with his logical entropy, one can define a  conditional version of graphtropy.  If one has two graphs G and H over the same node-set U, one can define conditional graphtropy as

$$
h(G | H) = \frac{ | \textrm{git}(G \cap H)| }{\textrm{git}(H) }
$$

\noindent -- i.e. the odds, when choosing an edge (a distinction) in $H$, that one gets an edge that is also a  distinction in $G$.  Similarly one can define  (symmetric) graphtropic mutual information via

$$
m(G , H) = \frac{ |\textrm{git}(G \cap H)|}{ |U_G|^2}
$$

\subsubsection{Graphtropy of Combinations}

Next, and dropping the assumption that G and H have the same node-set, we can look at the categorical product $G \bigotimes H$, observing that

$$
h(G \bigotimes H) =\frac{|\textrm{git}(G) \textrm{git}(H)| }{|U_G U_H|^2} = h(G) h(H)
$$

The categorial product forms a  lattice if paired with the disjoint union $G \bigoplus H$; and we may calculate

$$
h(G \bigoplus H) = \frac{h(G) |U_G| + h(H) |U_H|}{|U_G| + |U_H|}
$$

\subsubsection{Graphtropy of Distributions}

Next, suppose that instead of two concrete graphs G and H, we have a  joint weight function $w(x,y)$ over the space of graphs $x \bigotimes y$ with some fixed node-sets $U_x$ and $U_y$.    Here $w(x,y)$ indicates the probability that a  random graph drawn from the specified distribution is of the form $x \bigotimes y$ for the $x$ and $y$ given.   

(Also, let us assume that the graphs G and H involved all have specific orderings to their node-lists.  This assumption will play a  meaningful role in some counting arguments below, when we get to ``graphtropic thermodynamics.'')

Note that $w(x,y)$ is not a  probability density function because the same graph $z$ may be represented as the categorial product of multiple different pairs $(x,y)$.   We can define a  corresponding density $p(x,y) = w(x,y) / n(x,y)$ where $n(x,y)$ indicates the number of other pairs $(x', y')$ with $(x',y') \neq (x,y)$ and $x \bigotimes $y$ = x' \bigotimes y'$.

We can then define the graphtropic mutual information of the distribution $p$ as:

$$
m(x,y) = h(x) + h(y) - h(x,y) 
$$

\noindent Here $h(x)$ represents the odds that $(a,a_1)$ is not in the specified graph $x$, and h(y) represents the odds that $(b,b_1)$ is not in the specified graph $y$.   $h(x,y)$ represents the odds that we have both $(a,a_1) \not\in E_x$ and $(b,b_1) \not\in E_y$.  If the weight $w$ is set up so that $x$ and $y$ are drawn independently, then $h(x,y) = h(x) h(y)$.   

In a  similar way we can define graphtropic conditional information, KL divergence, and Fisher metric.   The latter enables us to define a  natural information geometry on the space of distributions over graphs, which is interesting in the context of the semantic interpretations to be discussed below. 

\subsection{Semantic Interpretation of Graph Logical Entropy in Terms of Distinction Graphs}

The above mathematics is subject to many possible semantic interpretations; however, it was developed with a  particular interpretation in mind, involving graphs representing relationships between observations by cognitive systems.  (For now, we leave the innards of these ``cognitive systems'' out of consideration, but later on we will introduce the option of modeling these cognitive systems as Dynamic Distinction Graphs, bringing the observer underlying the posited observations more fully into the fold of the theory.)

In this interpretation, we consider nodes in a  graph as possible observed entities, observed by some system.   We can model a  system as receiving a  set of percepts, each one arriving during a  certain interval of time, an ``entity" then corresponds to a  set of percept-sets (each percept-set in the set, being an ``instance'' of the entity).  Then, a  link between two nodes indicates that the two entities are {\bf perceived as identical} -- i.e., non-distinguishable --by the given observer.

In the case that distinction (or equivalently non-distinction, i.e.``perception as identical'') is transitive, the graph consists of a  set of disconnected components, each of which is a  fully connected graph -- i.e. a  partition.  But in the case of a  real-world perceiving agent, non-distinction is generally not transitive; because, after all, perceived entities can continuously morph into quite different perceived entities.

A subgraph of linked entities may also be considered a  higher-level entity (formed of lower-level entities with distinguishability relations between them).

The reader may see where we're going with this.  In the case of transitive non-distinction, we are dealing with partitions and hence the concept of logical entropy applies.  In the case where non-distinction is intransitive, we are dealing with graphs that do not represent partitions, and graphtropy is the relevant concept for measuring information.

\subsubsection{Combinatory Operations on Distinction Subgraphs}

To combine entities $A$ and $B$ semantically, we can use the Boolean or categorial operator-sets.

For instance, suppose $A$ refers to blue chickens, and $B$ refers to red dogs.   Then,

\begin{itemize}
\item The categorial product $a   \bigotimes B$  is the set of pairings of a  blue chicken and a  red dog.   
\item The categorial sum $A \bigoplus B$ is the set of entities to be used in building pairings of a  blue chicken and a  red dog (where if there is some individual that is both a  blue chicken and a  red dog, it  gets double-counted).   
\item The Boolean product $A \cap B$ is the set of entities that are both blue chickens and red dogs.  
\item The Boolean sum $A \cup B$ is the set of entities that are either blue chickens or red dogs.
\end{itemize}

\subsection{Key Properties of Graphtropy}

In \cite{ellerman2013introduction}, Ellerman compares Shannon entropy and logical entropy on various points.  Similarly we may compare graphtropy on these points; the results are shown in Tables \ref{tab:graphtropy-features} and \ref{tab:graphtropy-prob-features}.

\begin{table}[]
\centering
\caption{Key features of graphtropy algebra}
\label{tab:graphtropy-features}
\begin{tabular}{ll}
Elements &  dits $(u,u')$ of G \\
Order  & Refinement, $dit(G) < dit(H)$ \\
 Top of Order & graph with no edges \\
 Bottom of Order & completely connected graph \\
 Variables in Formulas &  graphs on vertex set $V$  \\
 Operations &  graph operations (e.g. categorial product, disjoint union) \\
 Formula $\Phi(x,y,\ldots)$ holds & $(u,u')$ a  dit of $\Phi(G,H,\ldots)$ \\
 Valid formula & $\Phi(G,H,..) = 1$ for all $G, H,\ldots$
\end{tabular}
\end{table}

\begin{table}[]
\centering
\caption{Probabilistic interpretation of graphtropy algebra}
\label{tab:graphtropy-prob-features}
\begin{tabular}{ll}
 Outcomes & edges in graph \\
 Events & graphs with vertex set $v$ \\
Event occurs & $u, u'$ are not linked (are distinguished) \\
 Quantitative measure & $p(G) = |git(G)| / |VxV|$ \\
 Random drawing & Prob. that graph $G$ does not have the link randomly selected
\end{tabular}
\end{table}

\section{Probabilistic Graphtropy}

Extending the above ideas, we can also look at graphs with weighted links.  Here the semantic interpretation is a  ``weighted distinction graph": the link between the node representing entity a  and the node representing entity B, represents the probability that the cognitive agent in question can distinguish an instance of a  and an instance of B.

Many types of weights can be considered, e.g. individual probabilities or probability intervals.  To make the math comes out simpler, we will focus here on the case where the weights are interval probabilities, e.g. $[.6, .8]$.   These may be interpreted as imprecise probabilities a  la Walley \cite{walley2000towards}.  The case where weights are indefinite probabilities (a la \cite{PLN}) is an immediate extension.

In this case, instead of the probability of there being a  link between $A$ and $B$, we can look at the distribution of link weights between $A$ and $B$ (given that $A$ and $B$ are randomly drawn from $B$), and at parameters of this distribution, such as the mean.  For a  non-weighted graph, this distribution is bimodal, and its mean is the percentage of pairs of nodes that are interlinked.

We may define ``weighted graphtropy" as graphtropy in the context of weighted graphs.  Key features of the weighted graphtropy are shown in Tables \ref{tab:weighted-graphtropy-features} and \ref{tab:weighted-graphtropy-prob-features}.  Here we define an order between two edges in a  weighted graph via $(u,u',w) < (u, u', w_1)$ iff $w<w_1$, i.e. $w_L>{w_1}_L , w_U < {w_1}_U$; i.e. if the interval probability $w$ is more specific than the interval probability $w_1$, and the former is contained within the latter.

An ``event" in the context of a  weighted distinction graph corresponds to a  second-order probability distribution, i.e. a  statement regarding the probability values on the edges of the graph (which is also a  statement regarding the links existing in the graph, implicitly).

The value $|E|$, where $E$ is a  set of weighted edges, in general could be given by some probability density $p$ over graph space.  It would then denote the percentage of graphs $G$, according to $p$, so that: For each edge $e \in E$, there is some edge $e' \in G$ so that $e' < e$.

\begin{table}[]
\centering
\caption{Key features of weighted graphtropy algebra}
\label{tab:weighted-graphtropy-features}
\begin{tabular}{ll}
Elements &  dits $(u,u', w)$ of G , where $w$ is an imprecise probability\\
Order  & Refinement, $dit(G) < dit(H)$\\
 Top of Order & graph with no edges \\
 Bottom of Order & completely connected graph, all edges with weight 1 \\
 Variables in Formulas &  weighted graphs on vertex set $V$  \\
 Operations &  graph operations (e.g. categorial product, disjoint union) \\
 Formula $\Phi(x,y,\ldots)$ holds & $(u,u',w)$ is less than some dit of $\Phi(G,H,\ldots)$ \\
 Valid formula & $\Phi(G,H,..) = 1$ for all $G, H,\ldots$
\end{tabular}
\end{table}

\begin{table}[]
\centering
\caption{Probabilistic interpretation of weighted graphtropy algebra}
\label{tab:weighted-graphtropy-prob-features}
\begin{tabular}{ll}
 Outcomes & edges in graph \\
 Events & graphs with vertex set $v$ \\
Event occurs & $u, u'$ has weight within interval $w$ \\
 Quantitative measure & $p(G) = |git(G)| / |VxV|$ \\
 Random drawing & Prob. that graph $G$ does not have the link randomly selected, \\
  & with a  weight inside the interval $w$ randomly selected
\end{tabular}
\end{table}

\section{Graphtropic Thermodynamics}

One of the most interesting aspects of Shannon entropy is the bridge that it forms between thermodynamics and information theory.   Graphtropy turns out to form a  different, related sort of bridge between these disciplines.

\subsection{The Analogue of Thermodynamic Entropy on Distinction Graphs}

The thermodynamic entropy of a  macrostate consists of the log of the number of microstates consistent with that macrostate.
The Shannon entropy approximates (but does not equal exactly) the number of micro states consistent with the observed macro state (the observed probability distribution over partition cells).

What are the micro and macro states in the above interpretation of graphtropy.  Consider first the case of partition graphs.  Here a  graph node consists of a  microstate; and a  ``macrostate"  consists of a  set of statistical observations of the graph, where what is observed is the percentage of the observations that lie in each partition cell.

For a  distinction graph, then, the number of ``microstates'' consistent with the graph is the number of automorphisms of the graph.  If the graph is a  partition graph, then this yields the entropy according to standard calculations.

For a  weighted distinction graph $G$, we want to average, over all automorphisms $s$, the distance between $s(G)$ and $G$.  Here distance may be measured e.g. by the average over all edges of the KL divergence between the distribution given to $E$ by $G$ and the distribution given $E$ to by $s(G)$.

\subsection{The Analogue of the Maximum-Entropy Distribution on Distinction Graphs}

Suppose we have a  ``node-weighted graph'' -- i.e., each node has an integer weight on it, drawn from some fixed finite set of integers.    Suppose the sum (or, equivalently, average) of the weights is supposed to be kept equal to some constant $K$.   

Suppose we also have some constraints on the graph structure -- e.g. 

\begin{itemize}
\item that the graph has to be a  set of disconnected completely-connected graphs; or 
\item that the graph has to have maximum degree less than $m$; or
\item that the graphtropy must be equal to, or closely approximated by, some fixed value (we will return to this case below)
\end{itemize}

\noindent Then, what is the most likely way for the weights to be distributed among the nodes?

Suppose we adopt the principle that, given lack of any other information, we should assume the various microstates compatible with an observed macrostate to have equal probability.  In this case, the  most likely way for the weights to be distributed, is the one that can occur in the most ways.   

For each node-weighted graph $G$, we can calculate the number of automorphisms of the graph that preserve the weights on the nodes, as well as the connection structure.  Specifically, each automorphism has got to map a  node with weight $m$ into another node with weight $m$.   

The most likely weighted graph is then going to be the one that has the greatest number of node-weight-preserving automorphisms.  

This is the ``maximum likelihood distribution" on $G$.

For a  partition graph in which all nodes in the same partition cell $i$ have the same weight $n_i$ , this yields the ordinary maximum entropy distribution associated with the probabilities $p_i$ defined by the sizes of the partition, subject to the constraint $\sum_i n_i p_i = K$.

If $G$ has probabilistic weights on its links, we can make a  similar definition, by restricting attention to automorphisms that map each link with weight $w$ into another link with weight $w' < w$.   Or more adventurously, we can look at ``$\epsilon$-automorphisms" $\Phi$ for which: where $w(a)$ is the weight on node $a$ and $\mu_{ab}$ is the mean of the imprecise probability weight on the link between $a$ and $b$, and some specific assumed small numbers $\epsilon_i$,

\begin{eqnarray*}
| w_a - w_{\Phi(a)}| < \epsilon_1 \\
1 - \epsilon_2 < \frac{\mu_{ab}}{  \mu_{\Phi(a) \Phi(ab)} } 1 + \epsilon_2\\ 
\mu_{ab} \mu_{\Phi(a) \Phi(ab)} ( | w_a - w_{\Phi(a)}| + |w_b - w_{\Phi(b)}| ) < \epsilon_3
\end{eqnarray*}

\noindent for all nodes $a$, $b$.   This is a  sort of ``fuzzy automorphism" -- the weight on the image $\Phi(a)$ is allowed to differ a  bit from that of $a$, and the weight on the link $(\Phi(a), \Phi(b))$ is allowed to differ a  bit from that of $(a,b)$.  But the differences can't get too big, and specifically, it matters more if strongly connected nodes are mapped into strongly connected nodes, than if weakly connected nodes are mapped into weakly connected nodes.

For partition graphs and linear constraints on the node weights, we know the maximum likelihood distribution is given by the Gibbs distribution.  Whether there is an equally elegant formula for more general distinction graphs is not clear.  There is some combinatorial work to be done here.  Given a  certain fixed graphtropy $g$, and $m$ integer labels (say, $1 \ldots m$) constrained to have a  fixed average, is there a  closed-form formula for he most likely distribution of labels (given constraint that two linked nodes must have the same label)?  A  closed-form formula for the $\epsilon$-automorphism case is less likely, but this case is perhaps more often directly relevant in practice.

\section{Connecting Graphtropy with Algorithmic Information}

There are well known and interesting connections between the Shannon information and the algorithmic information (the Kolmogorov complexity) \cite{teixeira2010entropy} \cite{zurek1989algorithmic}.   There turn out to be different, also interesting connections between the graphtropy and the algorithmic information.

Suppose the states represented by the nodes in the distinction graph $G$ are programs running on some universal computer $U$ (such programs may be represented, for example, as bit strings; though we will explore a little later the possibility of expressing such programs natively within distinction graphs by adding extra link types)   Suppose the observer, who is making the distinctions represented in the graph, has a  fixed algorithmic information content $K$.   Then, consider two nodes $N_1$ and $N_2$ with algorithmic information content $M>>K$.   The odds that the observer can distinguish $N_1$ and $N_2$ based on their contents decrease as $M$ increases.   There are roughly $2^M$ possibilities with algorithmic information content $= M$, but the observer can only distinguish $2^K$ at most; so the odds that observer can distinguish two random nodes with algorithmic information $M$ is around $2^{K-M-1}$ (the odds that $N_1$ and $N_2$ lie in the same one of the $2^{K-M}$ partition cells).

Thus the probability of a  link between $N_1$ and $N_2$ is $1-2^{K-M-1}$ -- a  direct relation between algorithmic information and graphtropy.

To sum up our conclusion: If the nodes in a  graph represent randomly chosen states with algorithmic information  $M$, and the observer's algorithmic information is $K$, then the graphtropy is $1-2^{K-M-1}$.

The graphtropy of a  distinction graph, constructed relative to an observer, is therefore considerable as a  measure of how much excessive algorithmic information exists in the system of observations modeled by the distinction graph, relative to the observer.   Or to put it more simply, {\bf the graphtropy measures how much more complexity there is in the environment relative to the observer.}

If the nodes represent randomly chosen states with algorithmic info $< M$, then a  similar formula can be derived, and the same conceptual conclusion holds.

\subsection{Graphtropic Energy}

There is also a  connection between these ideas and Tadaki's \cite{tadaki2008statistical} \footnote{see also \cite{baez2012algorithmic} for related considerations} formulation of the connection between algorithmic information theory and statistical mechanics.   Recall that in our discussion of maximum entropy distributions above, we assumed each node in our distinction graph was weighted with a  non-negative integer.   If these integers $v_i$ are taken to represent the algorithmic information content of a  bit string or other state resident at each node $i$, then we can interpret the sum

$$
\frac{ \sum_{i=1}^N v_i}{N}
$$

\noindent as a  kind of ``graph energy," where the possible values taken by the algorithmic information play the role of energy eigenstates in statistical mechanics.    If the possible values for the $v_i$ are $E_j$, then the graph energy may be rewritten

$$
\sum_j E_j p(E_j)
$$

\noindent where $p(E_j)$ denotes the percentage of nodes in the graph with $v_i=E_j$.

Given a  graph of this nature, in which each node holds a  state possessing a  certain amount of algorithmic information, {\bf the graph energy corresponds to the average algorithmic information across the nodes of the graph}.   It may be interesting to explore the maximum entropy distribution according to a  constraint of fixed average algorithmic information, and under various assumptions regarding the distribution from which the distinction graph is drawn.

If the graph is a  partition graph, then the maximum entropy distribution takes the form of the standard Gibbs distribution, and the energy takes the form of the ``algorithmic energy" presented in \cite{tadaki2008statistical}.   

\subsection{Graphtropic Temperature}

An interesting feature of the analysis in \cite{tadaki2008statistical} is that the compression ratio (the maximum compression ratio achievable by any computable program) is shown to be a  formal analogue of the temperature in thermodynamics: the more compressible, the lower the temperature.   For the Gibbs distribution in thermodynamics, the temperature plays a  role as a  parameter determining the variance of the distribution: higher temperature means less concentration of the distribution in the most probable states.   It is natural to ask whether the role of the algorithmic temperature is similar in the case of distinction graphs.  This seems likely, although validating this rigorously would require some graph combinatorics that we will not unfold here.

If we view nodes of the distinction graph as labeled with programs of fixed length $S$, then the compression ratio determines the connection probability.  The more compressible the program strings are, the greater the connection probability; where, according to the logic sketched above, connection probability would depend exponentially on compression ratio.   According to the logic outlined above, if the nodes in a  graph represent randomly chosen states with size $S$ and compressibility $D$, and the observer's algorithmic information is $K$, then the graphtropy is $1-2^{K-\frac{S}{D}-1}= 1 - 2^{K-1} 2^{\frac{S}{D}}$.  Temperature guiding the concentration of the maximum likelihood distribution in its most probable states, then means the same thing as graphtropy guiding the concentration in this way.

Specifically, suppose we have $m$ (e.g. $m=\frac{S}{T}$ different values for the integer weights on the $N$ nodes (which may represent e.g. algorithmic informations of the states at the nodes); and suppose we have a  much large number of nodes than values $(N>>m)$.  Then it seems intuitive that the maximum-likelihood distribution is going to be at its flattest when the connection probability is such that each node connects to around $\frac{m}{N}$ other nodes.    If so, this would mean that, roughly speaking, the maximum-likelihood distribution will tend to get flatter as the temperature descends toward  $\frac{m}{N}$ (which we have assumed is quite small).   As the connection probability goes below $\frac{m}{N}$, strange things may happen; but at this point we are verging away from the statistical foundation of the concept of temperature, because the temperature is getting around the granularity of the weight values.   In any case, these issues seem somewhat subtle and will benefit from detailed calculations.

\section{Dynamic Distinction Graphs: a  Phenomenology-Based Computational Model}

We now show how to extend the basic construct of a  distinction graph to  that of a  {\bf dynamic distinction graph} (DDG), which incorporates a  small number of additional features sufficient to implement an elemental model of computing.   Unlike most computing models out there,  the DDG model is fundamentally inspired by phenomenology rather than physical computing machinery or abstract mathematics.  Philosophically, we can think of DDG as a model of ``cognitive computation'' -- i.e. it is a computational model whose primitives are natural mental operations.  We believe it may be found natural for modeling dynamical and computational phenomena at the border between classical and quantum physics, and phenomena at the border between physical and cognitive operations (e.g. measurement and consciousness).  

As in the above, we start with a  ?distinction graph? in which node $x$  is linked to node $y$  if the given observer can?t distinguish $x$  from $y$.   Here $x$ and $y$  here are categories of observations, made by some observer.

An asymmetric link from $x$ to $y$ indicates that if an observer is observing $x$, they can't notice $x$ changing into $y$.   A symmetric link from $x$ to $y$ indicates that both asymmetric links from $x$ to $y$ exist.

Then introduce degrees of distinction, where $x$  is linked asymmetrically to $y$  with weight $p$ if the given observer can distinguish $x$ changing into $y$  $(1-p)\%$ of the time.

Note that one can consider there to be an implicit time-scale here: sometimes an observer may notice $x$ changing into $y$ only if the change is sufficiently rapid, but not if the change occurs slowly.   So in calculating the probability $p$, some average over different ways of changing is implicitly assumed, including an average over rates of change.   The basic model of the DDG, however, does not depend on any specific assumptions made in this regard.

Next, introduce weights or counts for nodes, indicating how often observations classifiable as node $x$ are encountered in the observer?s experience.   These weights may obviously be combined with the weights on distinction-graph links, in standard probabilistic formulas such as Bayes' rule.

So far we have something very similar to the underlying semantic model of OpenCog's Probabilistic Logic Networks framework \cite{PLN}.   If all the specifics are defined consistently, then the degree of distinction between $x$ and $y$ will be a monotone (nonlinear) transformation of the strength of the PLN Inheritance link between $x$ and $y$.   Because the Inheritance link measures what percentage of $x$'s properties are shared by $y$, and clearly the larger this number is, the smaller the chance that an observer can distinguish $x$ changing into $y$.

To make the model dynamic, we next introduce the notion of increasing or decreasing degrees of distinction -- $x$ or $y$  may be ``''getting more distinct'' or ``getting less distinct.''

This enables us to introduce the notion of causal implication between links, e.g. links embodying relationships such as

\begin{itemize}
\item If $x$  and $y$  get more distinct, then $w$  and $z$   get more distinct
\item  If $x$  and $y$  get more distinct, then $w$  and $z$   get less distinct
\end{itemize}

These implication links may have weights as well (and these weights will also embody some assumptions about distribution over time-lags).

In the PLN framework, temporal and causal implication links may be tagged with specific distributions over lengths of time, so that different such links in the same graph may pertain to different time-distributions.   This can also be done within the DDG model as needed.

To get a general, abstract computation model out of this, we then let causal implication links point to other causal implication links, so e.g. we have things like

\begin{itemize}
\item If $x$  and $y$  get more distinct, then it becomes more strongly true that: If $w$  and $z$   get more distinct, $u$ and $v$ get less distinct?
\end{itemize}

Note that grouping is basically there already as part of this framework.  If one has a  set of nodes that are completely interconnected, then this set of nodes can be treated as a  whole within implications (because the nodes are indistinguishable).

The operation of replacing a  node with a  completely connected subgraph should also be considered a  primitive (and it?s important because doing this substitution sets the stage for the nodes in the subgraph to become less completely connected).

This is a simple computational model founded on a handful of basic phenomenological primitives:

\begin{itemize}
\item distinct
\item indistinct
\item increasing
\item decreasing
\item cause = temporally-imply
\item always / often / sometimes / rarely / never 
\end{itemize}

The logic corresponding to this computational model via a Curry-Howard style correspondence has not been worked out in detail, but would seem at first blush to be a simple flavor of higher-order predicate logic.

Given an observed time-series of distinction graphs, rules such as the above can be data-mined using probabilistic causal inference methods, and a DDG model of the time-series can be reconstructed.   This DDG can then be operated going forward, as a simulation model corresponding to the observed time-series.

A node $x$ in a DDG, plus the causal implication links associated with the node, may be considered as a computational element; and its complexity may be assessed simply by counting the number of causal implication links involved (including links pointed to by higher-order links involving $x$, and so forth recursively).   The algorithmic information of a node in a distinction graph $G$ then consists of the minimal number of causal implication links needed to give the node $x$ the dynamics that it has in the graph $G$.  The correspondences between algorithmic information and graphtropy touched earlier can then be explored, with all the computation-theoretic and information-theoretic action occurring within the same distinction graphs enhanced with causal implication links.

\subsection{Modeling Observers as Dynamic Distinction Graphs}

DDGs allow us to bring the graphtropy model full circle, via dynamically modeling an {\it observer} as a DDG, aka an enhanced distinction graph.   An observer $A$'s action of observation may be modeled by considering certain of $A$'s DDG's nodes as ``sensors'', or by considering each of its DDG nodes as ``sensory'' to a certain degree, some more than others.   From the perspective of a second observer $B$, which divides possible objects of sensation into categories, one can then calculate whether $A$ reacts the same way (in terms of the dynamics within $A$'s DDG on exposure to sensations) to sensations in category $x$ and category $y$, and create the link between $x$ and $y$ in $A$'s distinction graph.  

Similarly, $B$'s division of objects of sensation into categories may be understood from the view of a third observer, $C$, who looks at the state of $B$'s DDG upon exposure to these sensations, and notes if there are differences between sensations in category $x$ and category $y$, etc.   One does not escape the ultimate observer-dependence of all distinctions.  But one can push it higher and higher up the hierarchy, modeling observers, observers of observers, etc. using a simple model of weighted distinctions associated with causal implications.

\section{The Second Law of Thermodynamics on Distinction Graphs}

The distinction-graph approach provides a  new view of the law of entropy non-decrease, making clearer the assumptions on which it rests as regards the observer's cognitive structure and relationship to the world.

Let us begin with the Second Law as formulated in terms of partitions.  We will move to more general distinction graphs a  little later.

As Baranger \cite{baranger2000chaos} and others have argued, the standard Second Law of Thermodynamics is rooted in the simple fact that, in a  complex dynamical system, it will often happen that two states in the same partition cell at time $t$ end up in different partition cells at time $t+1$.   Each such event increases the entropy of the overall partition; in terms of logical entropy it increases the number of dits.  

On the other hand,  one may also ask: If two states in different partition cells at time $t$ end up in the same partition cell at time $t+1$, does this not decrease the number of dits, the amount of entropy?

The key is to introduce the notion of an observer with a  memory.   Suppose we assume our observer has a  long-term memory comprising a  series of partitions, one at each point in time; and that the observer cross-indexes these partitions, so that it considers the same nodes to exist in the partitions associated with different times.   In general one can consider observers for whom the partitions at different times may involve overlapping but not identical node-sets; but for simplicity, let us now restrict attention to the case where the partitions at different times involve the same node-set.

Let us associate a  ``memory meta-partition'' with the observer: in the meta-partition, two nodes go in different partition cells if they have {\it ever} (in the observer's memory) gone in different partition cells.   That is to say, the observer's memory meta-partition is the lattice join of the time-specific partitions in the observer's memory.

In this case:

\begin{itemize}
\item When two nodes in the same cell at time $t$ end up in different cells at time $t+1$, they are then in different cells in the memory meta-partition forever after, so the number of dits in the memory meta-partition has increased
\item When two nodes in different cells at time $t$ end up in the same cells at time $t+1$, they remain in different cells in the memory meta-partition; so the number of dits in the memory meta-partition has not decreased
\end{itemize}

\noindent Clearly, we then have a  ``Second Law of Thermodynamics'' for the meta-partition: the entropy will never decrease and will often increase.

A similar argument holds if we forget about partitions and just look at distinction graphs.  If we introduce a  memory graph that consists of the join of a  series of historical distinction graphs, we will find that this join loses but does not accumulate links.   So {\bf the graphtropy of the observer's memory graph will never decrease}, and based on the observer's experience will generally eventually increase.   This is a  graphtropic version of the Second Law.

Of course, this conclusion will not hold exactly for a  real observer, because a  real observer will forget things, and this forgetting will cause previously noted distinctions to disappear in its memory graph.  Also a  real observer can develop and get modified in profound ways which result in introducing new nodes into its memory graph.   But the conclusion is nevertheless interesting.

This perspective on the Second Law makes clear that it relies on an assumption about the operation of memory.  Suppose our observer had precognition and thus, in its memory meta-partition, marked two nodes as distinguishable if they {\it ever}, in past or future, were distinguishable.  In this case, the memory meta-partition would not change over time and there would be no Second Law.  However, if its precognition were weaker or more erroneous than its ordinary memory, then one would still have a  probabilistic Second Law.

\subsection{Object Persistence and the Second Law}

It is interesting to dig a  little deeper here.  a  first step is to look at what happens if we have probabilistic distinctions and weighted distinction graphs; and thus a  probabilistically weighted memory graph.  In this case, to get the Second Law according to the above argument, what we need is some sort of auxiliary assumption such as

\begin{itemize}
\item Each time two nodes are viewed as distinct, this decreases the weight on their link in the weighted memory distinction graph.  
\item But when two nodes are viewed as the same, this does not increase the weight on their link as much; because it is assumed that their existing differences are simply not being observed at that time.
\end{itemize}

\noindent But this starts to seem a  little suspicious.  Why is it, actually that we assume previously observed distinctions are permanent, but previous observed non-distinctions are temporary?

To understand this better we need to ask: What does it mean to identify ``nodes'' observed at different points in time, in one's memory graph?   

To say that a  node $a^t$ observed during interval $t$ and a  node $b^s$ observed during interval $s$ represent the same entity, is presumably to say that there are observed patterns of interrelationship (with other nodes) that are common to $a^t$ and $b^s$.   We could represent these interrelationships in a  graph framework, e.g. by introducing hyperlinks between nodes to represent relationships of the form $e*f=h$ where $*$ is a  combinatory operator.  

For instance, if we are modeling nodes as harboring bit-strings encoding programs, $e*f$ might be interpreted as the output resultant from running the program at $e$ with the program at $f$ as input; saying $e*f=h$ then means that running $e$'s program with $f$'s program as input results in $h$'s program.  The identification of $a^t$ with $b^s$ is then a  statement that the two nodes are involved in similar sets of ternary links; where, two ternary links are judged as similar if they point between similar nodes (where nodes harboring the same program count as similar, providing a  base case for the recursive definition).   

In the Dynamic Distinction Graph framework introduced above, the ``ternary links'' here are simply causal implication links.   So one can consider two nodes in a DDG as similar if they are involved in similar causal implication links.   Similarity of two causal implication links may then be gauged partly in terms of similarity of the nodes they point to, which is a recursion, but not a vicious one; it's the kind of recursion that PLN deals with routinely.  

In this framework, the (graphtropic) Second Law then (speaking just a  bit loosely) appears to ensue from the assumption that 

\begin{itemize}
\item observing a  ternary link $e*f=h$ at a  certain time , increases the odds that another similar ternary link will be observed in future
\item the fact of  {\it not} observing a  ternary link $e*f=h$ at a  certain time,  increases the odds that no link similar to this one will be observed in future -- but by a smaller amount
\end{itemize}

\noindent This assumption appears to follow if one assumes that

\begin{itemize}
\item The observer's window into the world during any interval of time is quite incomplete (so that not seeing something at some moment, is not a  good indicate that said thing does not exist)
\item The world is not utterly random but has some continuity over time -- so that the relationships existing at one point in time have a  better odds of existing in the future, as compared to comparable randomly selected sets of relationships
\end{itemize}

\noindent These both seem quite reasonable assumptions in a  commonsensical context.   The second assumption is related to what Peirce called the ``tendency to take habits'' \cite{peirce1891architecture} and Lee Smolin called the ``precedence principle'' \cite{smolin2012precedence}.  In a   basic physics context it seems to ensue from such things as special relativity and the conservation of momentum and energy, which tend to keep many systems stable over time and prevent the world from looking like a  random succession of utterly disconnected situations.

What one gets from boiling down the Second Law to a  very basic level of observations and distinctions, is a  clearer understanding of the cognitive assumptions on which it rests.

\subsection{Graphtropy, States of Consciousness and Intelligence}

These reflections on the Second Law evoke a  connection between graphtropy and consciousness theory, which is in turn somewhat revelatory regarding the nature of intelligence, admittedly in on a speculative level.

From the perspective of consciousness theory, it seems that graphtropy is a  measure of the ``individuation'' associated with a  state of consciousness.   Minimum graphtropy occurs when everything is perceived as identical to everything else (an ``oceanic state" of total ``oneness" \cite{werman1986nature}); and maximum graphtropy occurs when everything is viewed as distinct from everything else.   

Real-world intelligent systems are necessarily somewhere between these extremes.  Seeing everything as identical will not lead to interesting complex adaptive behaviors with high probability.  On the other hand, real environments have too much complexity for any realistic bounded-resources intelligence to make all possible distinctions.   So real-world-intelligent systems will display intermediate graphtropy, in their internal cognitive distinction graphs.   This corresponds to our ordinary states of consciousness in which each experienced entity is merged into other highly (subjectively) similar entities, but not directly into all other entities.

The heuristic derivation of the graphtropic Second Law given above, depended on the assumption that the observer's memory graph was the join of his past memory graphs.  But of course, this is not realistic in practice, since any real-world observer is going to have a  memory that is very limited in size compared to the total sum of its historical perceptions.   In practice some past distinctions are going to be forgotten.   This suggests that the Second Law need not hold for real-world cognitive systems -- i.e. the graphtropy of a  real-world intelligent system's distinction graph need not steadily increase ... it may go up and down as the system learns and forgets.

Forgetting gives rise to the need for complex reasoning.  The practical inability to store and recall all of one's past particular distinctions, leads a  system to generate abstractions, which can then be used to imperfectly and probabilistically reconstitute past distinctions and to imperfectly and probabilistically apply the lessons of the totality of past distinctions to the future.   Intelligence is a  way of coping with the same memory imperfections that violate the assumptions leading to derivation of the Second Law.   Intelligence has to do with the regulation of the ongoing gain and loss of distinctions that keeps the graphtropy of a  complex adaptive system, most of the time, within reasonable homeostatic bounds, and in states of consciousness that are neither purely individuated nor fully oceanic.

These observations are not in themselves especially informative about the detailed aspects of real-world intelligent systems.   From the high-level perspective given here, about all we can say in this regard is that specific environments will tend to lead to the survival of systems with specific patterns of remembering and abstracting -- which is rather obvious and uninteresting.  The concepts given here do connect naturally with the train of thought presented in \cite{goertzel2009embodied}, in which the structure and dynamics of human-like minds is viewed as emerging from the need to achieve complex goals involving embodied communication, in environments heavily populated by other embodied communicating agents.   But for the purposes of the present paper, it is sufficient to observe that our treatment of graphtropy and its relation to observation in a  physics context, does fit naturally and sensibly with models of cognition as a  complex adaptive self-organizing process, and with models of the states of consciousness associated with cognitive systems.

\subsection{Maximum Entropy Production}

Another intriguing ``thermodynamic law" related to entropy is the ``maximum entropy production principle,'' which states that, in many cases, a  complex system will maximize the rate of entropy production.   The precise conditions under which this principle holds appear complex and are still under investigation from a  variety of perspectives \cite{martyushev2006maximum}.   The graphtropy perspective provides a  different, in some ways simpler and more revealing view.

``Graphtropy production'' may be defined as the rate of increase of graphtropy.  Consider a  time series of distinction graphs, with the same nodes but possibly different links at different times.  In a  non-weighted distinction graph, then, the graphtropy production is simply the number of links removed per unit time.  In a  weighted distinction graph, it is the average decrease of link weight per unit time.

Going back to the simple model introduced above in the analysis of the Second Law, we see that what determines the rate of graphtropy production, is the rate of identification of new ternary relationships that differentiate nodes $a$ and $b$ -- i.e. so that $a$ and $b$ would have appeared less differentiated if not for the new ternary relationships.   The more differentiations that are discovered and recorded in the memory graph, the more graphtropy.

To see why, under reasonable assumptions, the rate of graphtropy production might be maximized, let us return to the logic of partitions.   A  property of nodes, such as for example 

$$
P_a^2(c) = \exists a  \in N_G : a*x=c
$$

\noindent can be viewed as a  partition of the set of nodes into two categories: those nodes for which the property holds, and those for which it doesn't.  a  fuzzy property can be viewed as a  fuzzy partition also; but let's stick with strict properties for right now.

So, suppose a  system experiences a  number of events over a  certain interval of time, and that each of these events can be represented as a  ternary link in the manner described above.   Each ternary link $a*b=c$ defines six properties in a  natural way: the property $P_a$ described above has a  reverse

$$
P_c^{2*}(a) = \exists a  \in N_G : a*x=c
$$

\noindent and then there are the two properties

\begin{eqnarray*}
P_a^3(b) = \exists a  \in N_G : a*b=x \\
P_b^1(c) = \exists a  \in N_G : x*b=c \\
\end{eqnarray*}

\noindent and their reverses $P_b^{3*}(a)$, $P_c^{1*}(b)$.

So, each event that occurs creates 6 binary divisions of the nodes, according to each of these 6 properties.   If there are $M$ events altogether, then we have at maximum $6M$ binary divisions of the nodes, which in general divides the set of nodes into $2^{6M}$ partition cells, based on combinations of the properties implied by the events.

Now suppose we consider a  $2^{6M}$ dimensional probability vector, with a  probability $p_i$ for each cell.   If the number of nodes in the graph satisfies $N >> 2^{6M}$, then we have a  statistical-thermodynamics type situation.  Lacking additional constraints, the probability vector corresponding to the greatest number of microstates is going to be the one that makes each of these partition cells roughly equally numerous.

If $N$ is not quite big enough for this, we can look at binary or ternary combinations of properties only; for instance, looking at binary combinations only gives us a  $(6M)^2$ dimensional probability vector instead, which is much more reasonable; $N > (6M)^2$ looks more like an assumption that can commonly be fulfilled in practice.

In short, then: If we assume a  system is having a  bunch of random interactions during an interval of time, the most likely case is the one in which the Shannon entropy of the set of distinctions created by the interactions (from the perspective of an observer who sees all the interactions) is maximal.   This is also of course the set of distinctions for which the logical entropy is maximal.

Next, if we have an observer watching all these interactions over time, and remembering them, we will have a  situation in which the entropy of the system from the perspective of the observer is increasing very rapidly, because new distinctions keep getting added.  In other words, we see maximum entropy production.

The modeling of events in terms of ternary interactions on a  graph is not natural for all real-world situations (though it is adequate to model all computable dynamical systems and, under some interpretations, others as well).   However, we have introduced this specific model here mostly as a  simple way to make a  conceptual point.  The key concept here is that: {\bf From the point of view of an observer watching a  complex system change, the observed system will often demonstrate maximum entropy production because: The most likely scenario is for the events the system experiences to implicitly keep giving rise to system properties that overlap in a  maximally entropic way.}

The argument becomes more tedious to state, but the same conclusion appears to hold for the case of graphtropy production on intransitive distinction graphs.

Of course, if we take the ternary-interaction model of events, we conclude that eventually all possible properties of the 6 types described above (defined relative to nodes in the graph) will have occurred, so that the entropy will stop increasing.  But this takes us out of the ``statistical'' scenario where a  rule like maximum entropy production makes sense.  In the ternary-interaction model, one must consider the ``environment" of a  system as part of a  larger system; so that what is thought of as a  ``large heat bath'' in classical thermodynamics in this perspective just takes the form of more and more nodes in the distinction graph.

\section{Quantum Graphtropy}

Ellerman has extended the logical entropy to the quantum case, defining the ``quantum logical entropy''  of a  mixed state essentially as the degree of impurity of the corresponding density matrix \cite{ellerman2016classical}.  One can extend graphtropy to the quantum case in a  somewhat similar way, learning something about the semantics of quantum measurement in the process.

\subsection{Quantum Distinction Graphs as a  Representation of Mixed States}

Quantum logic, it is often said \cite{schlosshauer2013snapshot}, should be used by an observer to reason about things that this observer cannot in principle observe.  It follows from this that: Quantum logic is how an observer should reason about potential distinctions between things that it cannot distinguish from each other.  However, non-distinguishability by an observer is not necessarily a  transitive property of potential distinctions.  Speculatively, we suggest that perhaps weak quantum coherence may be effectively modeled in terms of distinction graphs where distinction is nontransitive in regard to the relevant observers. \cite{rieper2011quantum} \cite{li2012witnessing}. 

If $N_1$ and $N_2$ are linked in a  distinction graph defined relative to a  certain observer, then this observer should reason about their relationship using quantum logic.   What if $N_1$ and $N_2$ are linked, and $N_2$ and $N_3$ are non-linked, but $N_1$ and $N_3$ are not?   Then the quantum conclusion about $N_1$ and $N_3$ obtained from reasoning about their relation with $N_2$, has got to be consistent with the non-quantum conclusion obtained from reasoning about $N_1$ and $N_3$ based on direct observation. 

If the nodes in the graph represent quantum histories, so that the graph represents a  ``mixed quantum history", then the graphtropy represents the probability of getting different results in two independent  measurements with the same observable and the same pure state.   That is, graphtropy is a  measure of impurity in a  set of complexly related observables.

The full apparatus of graphtropy outlined above then applies to the case of a  graph whose nodes represent quantum histories; one ``merely" has a  different interpretation.

\subsection{Quangraphtropy}

John Baez and Blake Pollard have introduced the notion of {\bf quantropy} \cite{baez2015quantropy}, a  complex-variable version of the entropy, and have shown that the quantropy connects to basic quantum mechanics in a  similar way to how Shannon entropy connects to classical thermodynamics.   Using overlapping mathematics, a  complex-variable graphtropy (``quangraphtropy") can be connected to quantum mechanics in a  conceptually suggestive way.

Suppose we have a  complex number (the action) associated with each node on the graph $G$.   Suppose these are normalized so that their sum across the graph is 1; and suppose we have a  constraint on the expected action,

$$
\sum_{x \in N_G} A(x) a(x) = \langle a  \rangle
$$

What we can then ask is: given the distinguishability relationships between the states in the graph, what is the most likely graph consistent with the constraints?  This will be the graph that, consistent with the constraints, has the greatest number of automorphisms (where we are concerned only with automorphisms that  map a  node $x$ with action $a(x)$ into another node with $a(x) = a(y)$).

In a  simple situation (a large partition graph), this will come out the same as in quantropy formulation of quantum theory.

But what if the graph is not a  partition graph?   Then the most likely graph will be more complex, representing a  set of non-transitive "distinguishability" relationships between observable entities.   Since in the quantum context, non-distinguishability implies entanglement, the most likely situation will then be a  complex web in which various entities are non-transitively entangled with each other; i.e. with respect to a  particular observer, A may be significantly entangled with B, and $y$  significantly entangled with C, but a  not significantly entangled with C.   

This sort of intransitivity is not that odd of a  phenomenon; entanglement is a  type of correlation, and correlations are not generally transitive.   For instance, it can happen that several particles are entangled, yet if you ignore some  of the particles, the remaining pair is not entangled.  But these quantum distinction graphs constitute a  particularly simple way to model webs of intransitive entanglement.   It seems this may perhaps be an interesting way of modeling systems at the boundary between quantum and classical, i.e. systems displaying ``weak quantum coherence."

\subsection{Quantum Dynamic Distinction Graphs}

One can also articulate a quantum-computational version of a DDG, whose causal implication links involve; complex-number directional values, such as

\begin{itemize}
\item If $x$  and $y$  get more distinct in direction $a$, then $w$  and $z$   get more distinct in direction $b$ (by multiplier $\alpha$)
\item  If $x$  and $y$  get more distinct in direction $a$, then $w$  and $z$   get less distinct in direction $b$ (by multiplier $\alpha$)
\end{itemize}

\noindent It is not hard to see that enaction of such causal implication links is equivalent to multiplication of a complex number matrix by the complex number values associated with the nodes in the network.   If the nodes represent quantum histories, evolving according to a unitary matrix as happens in quantum mechanics, then the entries in the unitary matrix will imply the multipliers.   Conversely, a sufficiently complete set of such causal implication links will allow one to derive the unitary matrix driving the quantum system's evolution; or a less complete set of such causal implication links will imply a probability distribution over unitary quantum evolution matrices.

\section{Conclusion}

Beginning from a  very simple foundation (connection probability as a  kind of entropy) we have explored a  great variety of different concepts; some fairly straightforward and clearly-demonstrated, others more digressive and speculative.   For certain, more questions have been raised than answers provided.   Our hope is that graphtropy, as outlined here, will prove a  promising direction for diverse investigations.

While the math of graphtropy and distinction graphs is fascinating in itself, our primary motivation for exploring in this direction has been more philosophical and conceptual.   Given the intuition that ``information'' is a  foundational concept, both in physics and in psychology, one is drawn to try to understand, at bottom, {\it what information is}.   Communication theory, thermodynamics and quantum mechanics provide intriguing clues but appear not to get at the crux of the issue.   

The crux of the matter of information, we suggest, is {\it the body of distinctions made by a  particular observer}.   This body of distinctions may be modeled as a  graph, and may be quantified in various ways.   The graphtropy is one particularly interesting quantification, which reduces to more standard entropy notions (Ellerman's logical entropy, which relates with Shannon entropy in clear ways) under special assumptions.   Graphtropy relates naturally to phenomenological aspects of states of consciousness (e.g. degree of ``oceanic''-ness).  The graphtropy perspective also clarifies the semantics of quantum distinctions, via constructions like the ``quangraphtropy'' and quantum DDGs.   Exploring concepts like maximum entropy and maximum entropy production in the context of distinction graphs leads to some complex mathematics and unresolved particulars, but appears a  promising direction.   

Overall, we believe these preliminary investigations lend weight to the idea that, in order to understand the fundamental meaning of information and to better comprehend the application of information in various disciplines, one should carefully ground one's informational calculations and constructions in distinction graphs formed to represent the perspectives of particular observers.   Observers may also be modeled in terms of the distinctions they embody, and the causal relations between their embodied distinctions at various points in time, pointing toward a perspective in which a small number of primitive concepts (distinction, probability, time, causation) suffice to model the complex informational structures and dynamics of the physical and cognitive world.

\bibliographystyle{alpha}
\bibliography{bbm}

\end{document}